\documentclass[18pt]{report}
\usepackage[english]{babel}

\usepackage{fullpage}
\usepackage{titlesec}  

\usepackage{algorithm}
\usepackage{algcompatible}
\titleformat{\chapter}[display]
{\normalfont%
   \Large 
    \bfseries}{\chaptertitlename\ \thechapter}{18pt}{%
    }
\usepackage{amsmath,amssymb}

\usepackage{amsmath,amssymb}
\usepackage{hyperref}
\usepackage{wrapfig}
\usepackage{mwe}
\usepackage{lipsum}
\usepackage{acronym}
\numberwithin{equation}{section}
\usepackage{psfrag}
\usepackage{float}
\usepackage{times}
\usepackage[intoc]{nomencl}
\usepackage{physics}
\usepackage[utf8]{inputenc}
\usepackage{multicol}
\usepackage{psfrag} 
\usepackage{amsmath}

\renewcommand{\arraystretch}{1}
\usepackage[draft,enable-survey]{pdfpages}
\usepackage{pdfpages}
\usepackage{graphics}
\usepackage{csquotes}

\usepackage{geometry}
\usepackage[backend=biber,style=numeric,sorting=none]{biblatex}
\usepackage[thinc]{esdiff}
\usepackage{float}
\usepackage[utf8]{inputenc} 
\usepackage[T1]{fontenc}    
\usepackage{amssymb,amsmath,amsfonts} 
\usepackage{braket} 
\usepackage{caption}
\usepackage{algorithm}
\usepackage[noend]{algpseudocode}
\usepackage{gensymb}
\usepackage{libertine}
\usepackage{subfig}
\usepackage{wrapfig}

\begin{document}
\nocite{*}
\begin{titlepage}
    \begin{center}
        \Large\textsc{3D Reconstruction using Structured Light}\\
        \textsc{} \\
        \small\textbf{\textsl{A project report submitted to}} \\
        \small\textbf{\textsl{Visvesvaraya National Institute of Technology, Nagpur}} \\
        \small\textbf{\textsl{in partial fulfillment of the requirements for the award of}} \\
        \small\textbf{\textsl{the degree}} \\
        \textsc{} \\
        \Large\textbf{\text{BACHELOR OF TECHNOLOGY}}\\
        \Large\textbf{\text{IN}}\\
        \Large\textbf{\text{MECHANICAL ENGINEERING}}\\
        \textsc{} \\
        \textsl{by}\\
        \textsc{} \\
        \large{\textbf{Aman Gajendra Jain}\hspace{35pt} \textbf{BT16MEC100}}\\
        
        \textsc{} \\
        \large\text{under the guidance of} \\
        \textsc{} \\
        \large{\textbf{Dr. Shital Chiddarwar}}\\
        \textsc{} \\
        \begin{figure}[h]
            \centering
            \includegraphics{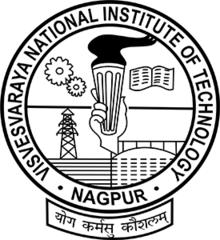}
        \end{figure}
        \Large{\textbf{Department of Mechanical Engineering}}\\
        \Large{\textbf{Visvesvaraya National Institute of Technology}}\\
        \Large{\textbf{Nagpur 440 010(India)}}\\
        \Large{\textbf{2020}}\\
        \small{\textcopyright} \small{Visvesvaraya National Institute of Technology (VNIT) 2020}
        
        \textsc{} \\
    \end{center}
\end{titlepage}

\pagenumbering{roman}

\begin{center}
\hspace*{\fill}{\Large\textbf{Department of Mechanical Engineering}}\hfill
\makebox[0cm][r]{\includegraphics[height=2cm]{vnit_logo.png}}%
\\\Large\textbf{Visvesvaraya National Institute of Technology, Nagpur}\\ 
\textbf{Declaration}
\end{center}
\large I, Aman Gajendra Jain, hereby declare that this project work titled "3D Reconstruction using Structured Light" is carried out by us in the Department of Mechanical Engineering of Visvesvaraya National Institute of Technology, Nagpur. This work is original and has not been submitted earlier whole or in part for the award of any degree/diploma at this or any other Institution/University.
\begin{table}[h!]
\large\text{Date: } \hspace{2cm}
\setlength{\tabcolsep}{5.5pt}
\renewcommand{\arraystretch}{1}
\begin{tabular}{|c|c|c|c|} 
 \hline
 Sr.No. & Enrollment No. & Names & Signature\\ 
 \hline
  1. & BT16EEE100 & Aman Gajendra Jain & \includegraphics[width=.2\textwidth]{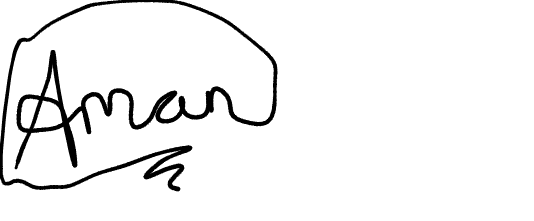}\\
 \hline
\end{tabular}
\end{table}

\begin{center}
    \textbf{Certificate}
\end{center}
\large This is to certify that the project titled \textbf{"3D Reconstruction using Structured Light"} submitted by \textit{Aman Gajendra Jain} in partial fulfillment of the requirements for the award of the degree of \textbf{Bachelor of Technology in Mechanical Engineering}, VNIT Nagpur. This work is comprehensive, complete and fit for final evaluation.

\begin{flushleft}
\vspace{0.75cm}
\hspace{290pt}\large\textbf{Dr. Shital Chiddarwar}\\
\hspace{303pt}\large{Associate Professor}\\
\hspace{270pt}\large{Dept. of Mechanical Engineering}\\
\hspace{314pt}\large{VNIT Nagpur}\\

\large{Head, Dept. of Mechanical Engineering}\\
\large\text{VNIT Nagpur}\\
\large\text{Date: }
\end{flushleft}

\chapter*{Acknowledgement}
\par First and foremost, we express our sincere gratitude to our guide, Dr. Shital Chiddarwar, Associate Professor, Department of Mechanical Engineering and Faculty-in-charge of IvLabs - the Robotics Club of VNIT, for her continuous support and guidance throughout the course of the project. We would also like to thank her for providing space to work in IvLabs. Finally, we would like to thank Saurabh Kemekar and Mayank Bumb for their help and useful discussions.

We thank Prof. Vilas R. Kalamkar, Head of Department, Mechanical Engineering for giving us the opportunity to proceed with this project. We thank the faculty of Mechanical Engineering Department for their teaching in our years at this institute, which has enabled us to take up this project.

\vspace{50pt}

\begin{center}
\includegraphics[width=.2\textwidth]{signature_1.png} \hspace{100pt} \\

\textbf{Aman Gajendra Jain} \hspace{100pt} \\
\vspace{50pt}

\end{center}

\chapter*{ABSTRACT}

\par The coordinate measuring machine(CMM) has been the benchmark of accuracy in measuring solid objects from nearly past 50 years or more. However with the advent of 3D scanning technology\cite{rocchini2001low}, the accuracy and the density of point cloud generated has taken over. In this project we not only compare the different algorithms that can be used in a 3D scanning software, but also create our own 3D scanner from off-the-shelf components like camera and projector. Our objective has been :

\begin{description}
  \item[$\bullet$] To develop a prototype for 3D scanner to achieve a system that performs at optimal accuracy over a wide typology of objects.
   \item[$\bullet$] To minimise the cost using off-the-shelf components
   \item[$\bullet$] To reach very close to the accuracy of CMM

\end{description}
\addcontentsline{toc}{chapter}{Abstract}

\listoffigures
\addcontentsline{toc}{chapter}{List of Figures}

\listoftables
\addcontentsline{toc}{chapter}{List of Tables}

\tableofcontents 

\cleardoublepage

\pagenumbering{arabic}

\chapter{Introduction}

\par The way humans perceives 3D information using binocular stereopsis has been the subject of research in computer-vision community for very long. They have tried to model stereopsis using two cameras as human eyes. This concept is known as stereo-vision and it is a field of active research.

\par The most common setup for stereo-vision is two cameras separated by a fixed horizontal distance, and having identical orientation, similar to human eyes, capturing the image of any arbitrary 3D scene simultaneously. From the captured images, it was observed that, the pixels of the objects, which were seen in both the cameras, were displaced more if they were nearer to the cameras and vice-versa. This pixel to pixel displacement is known as disparity and it is inversely proportional to the distance of object from the cameras. This is how eyes also perceive depth!

\par However, to extract depth information, we must have internal details of cameras like focal length,  principal point, distortions if any, etc; as well as the position relationship between two cameras. The process of extracting this information is known as camera calibration. Camera calibration will be discussed in detail in Chapter \ref{chapter:Camera_Calibration}. 

\par In order to get dense 3D reconstruction, we must be able to find for every pixel in image one, its corresponding pixel in the image two. This is known as correspondence problem in the literature. In order to solve for correspondence there are two well-known approaches :
\begin{itemize}
    \item Active Stereo-Vision Techniques
    \item Passive Stereo-Vision Techniques
\end{itemize}
   
\par Active techniques employs the use of additional light rays in the scene. In this techniques, a projector or a similar device, projects light rays onto a scene. Each projected ray of light encodes some information which is unique in some sense. Hence, owing to some structure encoded in light this is also known as structure light. This light rays are diffused by the objects present in that scene and are captured by the stereo-camera setup. After being captured as images, this unique information encoded in the scene is decoded for every pixel. This information is than used as index for matching. There are other similar methods in which instead of two, a single camera and a projector is used, This setup is known as structured light system. More details on structured light will be discussed in Chapter \ref{chapter:structured_light}. We might have to calibrate the projector in this technique and more on this will be discussed in Chapter \ref{chapter:projector}.

\par Passive techniques do not use any additional rays but use the geometric and photometric constraints to solve for correspondence problems. We will not discuss these techniques in detail.

After calibration and extracting correspondences it is essential to triangulate these correspondences to get back the 3D point cloud. This will be discussed in Chapter \ref{chapter:Triangulation}. For complex 3D object, point-cloud information from one view is not sufficient and hence we will be using a turn table based setup to capture multiple shots of the object. The mechanical assembly and the complete setup is discussed in Chapter \ref{chapter:Mechanical_Setup}. After getting point clouds from multiple views, it is essential to combine these point clouds. This is done using point-cloud registration algorithm discussed in detail in Chapter \ref{chapter:icp}.

This report is organized as follows. Chapter \ref{chapter:Mechanical_Setup} describes the mechanical assembly and camera projector setup in detail. Chapter \ref{chapter:Camera_Calibration} throws light on the camera calibration process. Chapter \ref{chapter:projector} discusses the projector calibration and camera-projector extrinsic calibration. Chapter \ref{chapter:structured_light} describes in details the different Structured Light algorithms with advantages and dis-advantages. Chapter \ref{chapter:Triangulation} explains the process of Triangulation. Chapter \ref{chapter:icp} describes the Point Cloud Stitching Algorithm. Chapter \ref{chapter:Results} describes our results.  
\chapter{Mechanical Setup}
\label{chapter:Mechanical_Setup}

The physical set-up consists of a projector and a camera as the two main components. In addition a projector stand to house the projector was constructed along with a turntable upon which the object to be scanned was placed. The projector stand provides a platform to place the projector and camera set-up. It’s slightly declined platform also enables us to get a wider view of the object to be scanned and not just from one plane. We get a view of the top plane and the front plane of the object to be scanned. If the object is then rotated we would get a complete view of the object. The turntable is constructed to serve just this function. 
\begin{figure}[htb!]
    \centering
    \includegraphics[width=\textwidth]{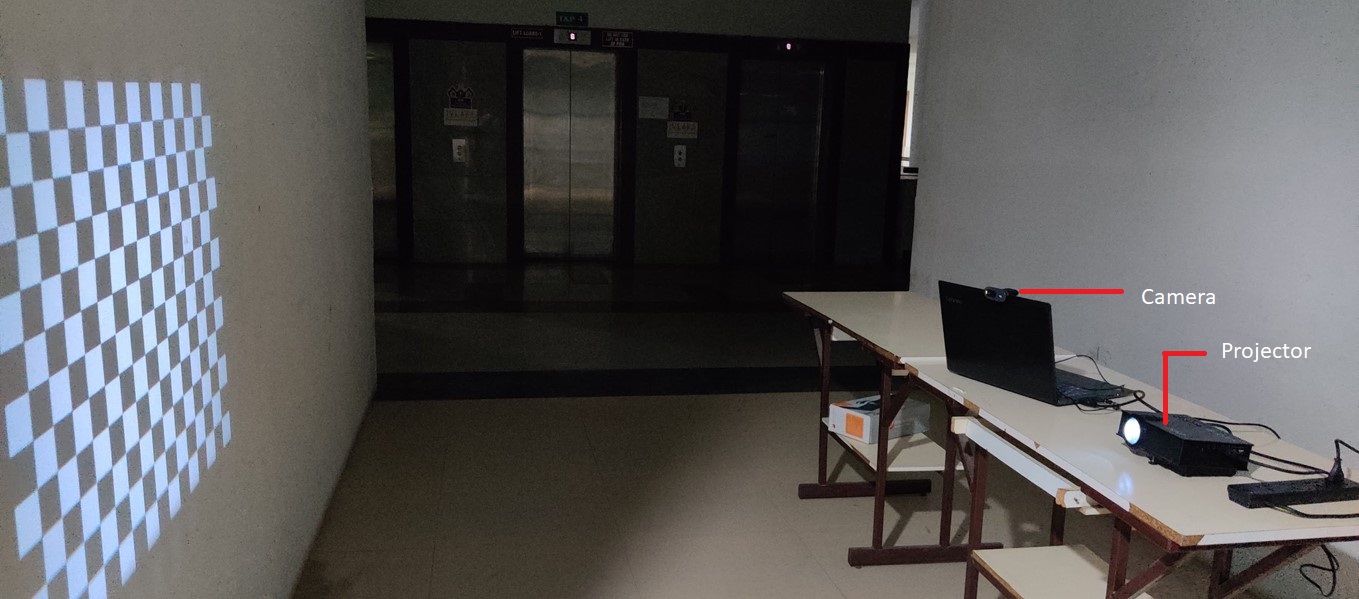}
    \caption{Camera Projector Structured Light System}
    \label{fig:SL system}
\end{figure}

The projector stand was constructed using L-beams and a composite plate. The composite plate is attached to the L-beams using simple bolts and nuts. The beams are about a metre in length which is adequate height for the required small scale application of this project. The plate can be adjusted to a suitable height by simply unscrewing the fasteners and sliding it to the required position. The plate is wide enough to accommodate the projector and two cameras along its length. The composite plate is also structurally capable of undertaking the load of the projector and the cameras while still being lightweight. The beams are equipped with drilled holes along its length so that the platform can be adjusted to a required height based on the dimensions of the object to be scanned. The beams have an L cross-section and are made of steel which keeps them very lightweight and are easily available in the market. The platform works along with the turntable. 
\begin{figure}[htb!]
    \centering
    \includegraphics{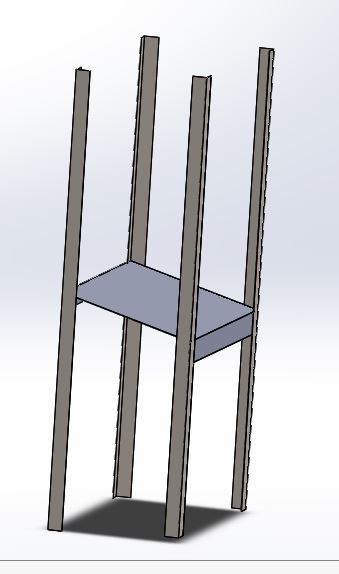}
    \caption{CAD model of Projector Stand}
    \label{fig:CAD stand}
\end{figure}

The turntable serves the function of rotating the object to be scanned so that its complete surface can be captured by the camera and the projector. It was constructed from plywood and steel brackets. These materials make the turntable very lightweight in construction. The turntable can also be easily dismantled and assembled using a simple wrench as it is put together using only nuts and bolts. It is square shaped and about 30 centimetres wide. The turntable does not structurally actually take the load of the object to be scanned. The load is exerted onto the motor which is attached to and situated below the rotating platform. In order to undertake larger loads a redesign would be required where the load would be taken by the rotating platform and turntable with the help of appropriate components such as ball bearings. But the current design is suitable for small scale applications and also lowers the cost of construction. 

The turntable is powered using a motor which is controlled through an Arduino microcontroller and powered with the help of an SMPS (Switched Mode Power Supply).  The motor is housed within the turntable and attached securely within a housing. The housing is constructed using simple steel brackets which securely hold it in place as it rotates the rotating platform of the turntable. The servo motor can be accurately controlled using the Arduino microcontroller and provide complete 360 degree rotation but one degree at a time. The rate of rotation or its speed can be adjusted to the required amount, accurately, using the Arduino microcontroller which is easily programmable and does not require previous experience in electronics or microcontroller programming. 

The motor used is a Dynamixel (MX-64) which is a servo motor and therefore uses PID control. The PID control or Proportional Integral Derivative control allows us to accurately control the process variables which angle of rotation of motor shaft, the rate at which it rotates and the power used, through control loop feedback mechanisms. The motor weighs 160 grams and is about 40 millimetres to 60 millimetres wide. Its shaft is equipped with a bracket which was used to attach it to the rotating platform of the turntable. The motor has a gear ratio of 200 to 1 which helps it generate torque. It can operate under a radial load of up to 40 Newton and an axial load of up to 20 Newton. It can reach rotation speeds of about 78 revolutions per minute at a rated voltage of 14.8 Volts. 

\begin{figure}[htb!]
    \centering
    \includegraphics[width=\textwidth]{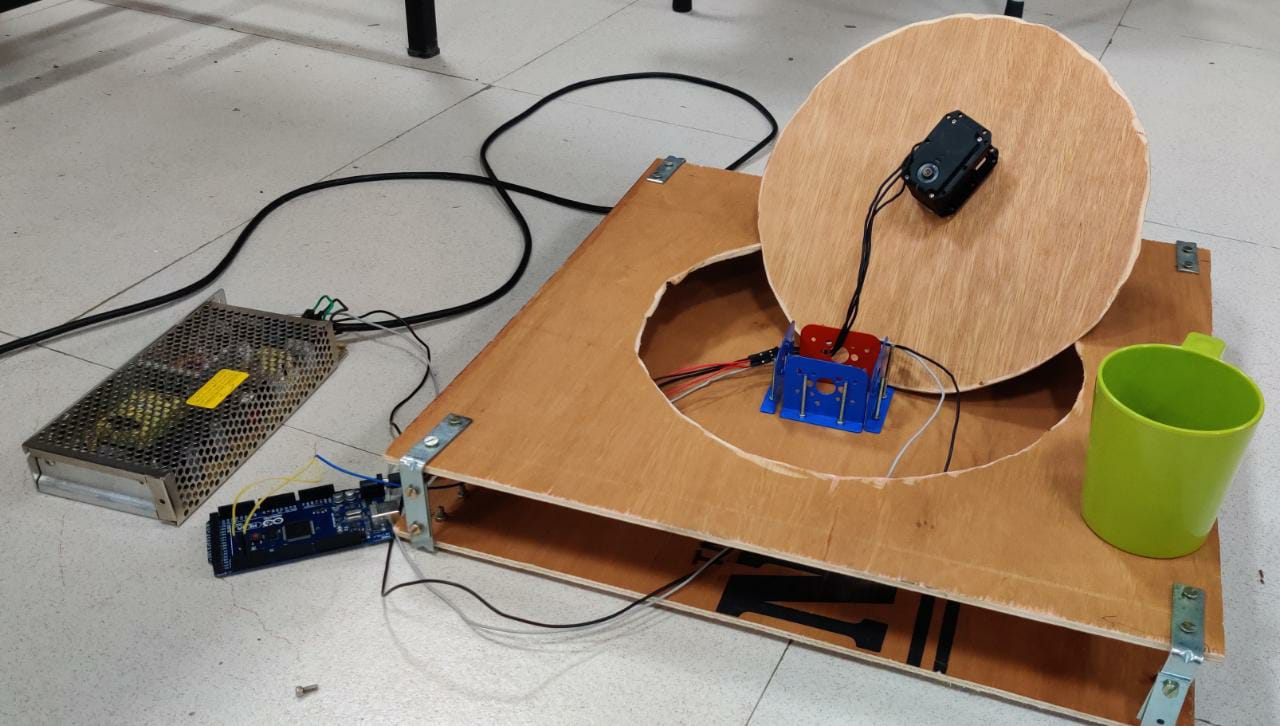}
    \caption{Turntable}
    \label{fig:turntable}
\end{figure}

\chapter{Camera Calibration}
\label{chapter:Camera_Calibration}
\section{Overview}
Camera Calibration also known as Camera Resectioning is the process of estimating internal parameters (also known as intrinsics) and external parameters (also known as extrinsics) of a camera. Internal parameter comprises of focal length, principal point, skew (if present), aspect ratio and lens distortions. External parameters comprises of 6 degrees of freedom associated with the camera in any arbitary coordinate system. In order to derive 3D(metric) information from a camera, calibration is an essential step. It allows photogrammetric measurements from images, distortion correction, 3D reconstruction, etc.

\section{Pin Hole Camera Model}
\begin{figure}
\includegraphics[width=\textwidth]{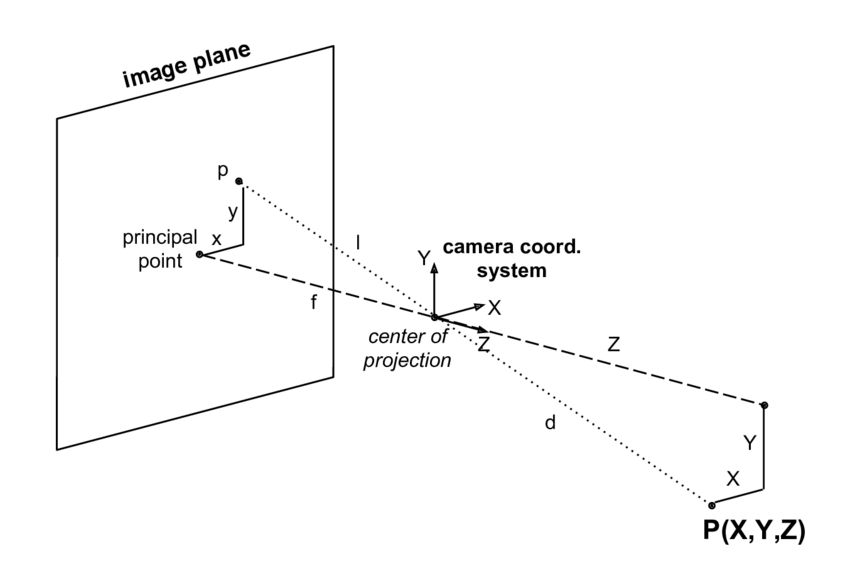}
\centering
\caption[Pinhole Camera]{Pinhole camera}
\label{fig:pinhole}
\end{figure}

In order to mathematically model digital cameras used in this project, it is important to gain understanding of a simple pinhole camera. A pinhole camera, unlike other cameras, do not have any lenses or mirrors, but just an aperture for light rays to enter and a film to capture the light. It is shown in the figure \ref{fig:pinhole}. 

The image obtained from pin-hole camera is inverted and is projection of 3D world object onto a camera film. The size of the projected image is dependent on the distance of object from pin-hole as well as the distance of pin-hole from the camera film (also referred as focal length).

The relationship between the 3D object and its 2D image for a pin-hole camera from figure \ref{fig:pinhole} can be expressed as :

\begin{equation}
\frac{y}{-f} = \frac{Y}{Z}  \ \textrm{or} \ y = -f * \frac{Y}{Z}
\end{equation}

Similarly,
\begin{equation}
\frac{x}{-f} = \frac{X}{Z}  \ \textrm{or} \ x = -f * \frac{X}{Z}
\end{equation}

Negative sign in the equations indicates that the images are inverted.
\subsection{Extending pin-hole camera model to digital cameras}
The pin-hole model can be extended to cameras by making few changes : - 
\begin{itemize}
    \item The origin of image lies on the top left corner, so we must shift the projected coordinates with (cx, cy), which represents the projection of camera center on the image plane plane
    \item In order to keep the equations positive 2.2.1 an 2.2.2 , we have assumed a virtual screen in-front of the camera, thus eliminating the negative signs.
    \item The image recorded is not continuous but discretized by the image sensor into numerous pixels. Each pixel should ideally be square. However, it deviates from a square, in terms of aspect ratio and skew.
    \item After accounting all the parameters, the new transformation from camera coordinate system to image in homogeneous coordinates can be written as follows :
    \begin{equation}
    \lambda
    \begin{bmatrix} u\\
    v\\
    1
    \end{bmatrix} = 
     \begin{bmatrix}
     f_x & s & c_x \\
     0 & f_y & c_y \\
     0 & 0 & 1
     \end{bmatrix}
     \begin{bmatrix}
     X\\
     Y\\
     Z\\
     \end{bmatrix}
    \end{equation}
    In the above equation 2.2.3, $\lambda$ is equal to Z, which is the distance of the world point from the principal point.
    \item The X, Y, Z in the equation 2.2.3, is expressed in camera coordinate system, in order to generalize it to any arbitrary Cartesian system we may introduce a Transformation Matrix T, from arbitrary coordinate system to camera coordinate system, thereby modifying equation 2.2.3 as :
    \begin{equation}
    \lambda
    \begin{bmatrix} u\\
    v\\
    1
    \end{bmatrix} = 
     \begin{bmatrix}
     f_x & s & c_x \\
     0 & f_y & c_y \\
     0 & 0 & 1
     \end{bmatrix}
     \begin{bmatrix}
     r_{11} & r_{12} & r_{13} & t_x \\
     r_{21} & r_{22} & r_{23} & t_y \\
     r_{31} & r_{32} & r_{33} & t_z
     \end{bmatrix}
     \begin{bmatrix}
     X\\
     Y\\
     Z\\
     1\\
     \end{bmatrix}
    \end{equation}
    The first three columns of transformation matrix constitutes a rotation matrix and the last column is the translation vector.
    \item The pinhole model deviates as we move farther away from the image center, this effect can be accounted using polynomial model of distortion. We can also incorporate tangential distortions if necessary. 
    \end{itemize}

\section{Zhang's Method of Camera Calibration}
In order to calibrate our camera, we have used Zhang's method \cite{791289, 888718} of camera calibration. This method requires us to capture images of an asymmetric checker board in different orientations and positions. The output of this method is a set of camera parameters : -
\begin{itemize}
    \item The intrinsic parameters like : -
    \begin{itemize}
        \item principal point ($c_x$, $c_y$)
        \item focal length in pixels ($f_x$, $f_y$)
        \item skew s (if present)
        \item distortion coefficients
    \end{itemize}
    \item The extrinsic parameters like : -
    \begin{itemize}
        \item The orientations of checkerboard w.r.t camera-coordinate system in different images
        \item The positions of checkerboard w.r.t camera-coordinate system in different images
    \end{itemize}
\end{itemize}

This information will be used in further processing.

\chapter{Structured Light}
\label{chapter:structured_light}
\section{Overview}
 Correspondence Problem is a well studied problem in computer vision literature. Attempts have been made to solve it using optical flow, block matching, neural networks, etc. Structured Light approach is one such technique that could very well resolve this problem. This technique is used in various 3D cameras like Intel Realsense, Microsoft Kinect, Orbec Astra and so many other RGBD cameras, to solve the correspondence problem in a stereo setup. 
 
 Before we begin with structured light it is essential to understand what is correspondence problem and why are attempts being made to solve the problem. So correspondence problem as the name suggest is a problem of finding correspondences between two or more images viewing the same 3D scene structure. Correspondence is the location of common 3D scene point in the image planes of two or more cameras. It is essential to find correspondences to perform various multi-view geometry tasks like visual odometry, 3D reconstruction, SLAM, Structure from Motion, calibration etc.    
 
 Now that we know what correspondence problem is and why it is essential, the next thing which needs to be understood is what is structured light and how does it alleviates the correspondence problem. Structured light is the process of projecting known patterns on to a scene under consideration. This known pattern encodes certain information which when decoded can help establish correspondence. To give an understanding of how correspondences are established, lets assume a pixel location or a blob in one image which captures structured light and encodes the string '1101101101', 
 so we search for this string in the next image, and at whichever location we find this string, that location or a blob region is a corresponding blob/pixel location to blob/pixel location in first image. What information or string the pattern encodes varies in terms of structured light used in the process. 
 
 So now that we are aware of what structured light is and how it alleviates the correspondence problem, its time that we study in details what are the different types of structured light and what information do they encode.

\section{Types of Structured Light}
All structured light techniques can be broadly classified into two categories :
\begin{itemize}
\item \textbf{Sequential or Multiple shots} - In this method, multiple patterns of structured light are projected on the scene and is/are captured by single/multiple cameras. This is generally used in 3D scanning when the object is static and no stringent constraint on capture time is imposed.
\item \textbf{Single shot} - In this method, only one pattern of structured light is projected and is/are captured by single/multiple cameras. This is generally used in a dynamic scene, or when there is a constraint on capture time.
\end{itemize}

For the purpose of 3D scanner, developed for scanning mechanical component it is the most appropriate category of structured light to be used. Since our object is stationary, and the constraint on scan time is also not very stringent.

The most commonly used sequential structured light used in 3D scanner are :
\begin{itemize}
    \item Gray coded Structured Light 
    \item Phase Shifted Structured Light
    \item Hybrid Structured Light
\end{itemize}

We will one by one study all the common sequential structured light patterns. Let us assume our setup consists of a camera capturing image of resolution 1280x720 and a projector projecting an image of resolution 1920x1080. In order to find unique correspondence between projector projection and image, we must have at-least 1920 x 1080 different strings for every row and column of the projector.

\subsection{Gray Coded Structured Light}
In gray-coded method of structured light projections, we project multiple images composed of horizontal and vertical black and white stripes arranged in a particular pattern\cite{Salvi04patterncodification, Inokuchi1984RangeImagingF3}. This pattern is decided by the information that is to be encoded. For instance, if we want to encode the x-coordinate or y-coordinate of projector pixel location, we will first identify the maximum length of string to be used in order to classify all 1920 values along the width of image and 1080 values along the height of image. The minimum bits required to do so would be $log_2(1920) \sim 11$ along width and $log_2(1080) \sim 11$ along height. It means we will need 11 images to encode every x-coordinate and 11 more images to encode every y-coordinate of projector pixel location using gray code.

Lets understand this with an example, suppose, we were to encode the pixel location (1152,648) of projector described above. We will first convert the decimal number \textbf{1152} to gray-coded string. It would be an 11-bit string - '\textbf{11011000000}'.  Now the encoding in the images would be done as follows, in the first image, all the values lying in column 1152 will become white as the first character in string is '1', similarly, in second image all the values lying in column 1152 will also become white because second character is again '1' but the in third image all the values lying in column 1152 will become black as the corresponding character in string is '0'. In this way for all 11 images corresponding column values would be decided by the gray-coded 11-bit string.  Similarly to encode the y-coordinate \textbf{648} with gray-code value '\textbf{01111001100}'. We will assign value black to row 648 in first image, assign white to row 648 in second image and so on for all 11 images. Here, I have performed encoding for only 1 pixel location, but we can simultaneously do this for all 1920 columns and 1080 rows. Thus we will obtain in total 22 images, 11  for decoding x-coordinate of projector pixel location and 11 more images for decoding y-coordinate of projector pixel location.

The gray-coded pattern projected on a scene looks like the image shown below : 

\begin{figure}
\includegraphics[width=.4\textwidth]{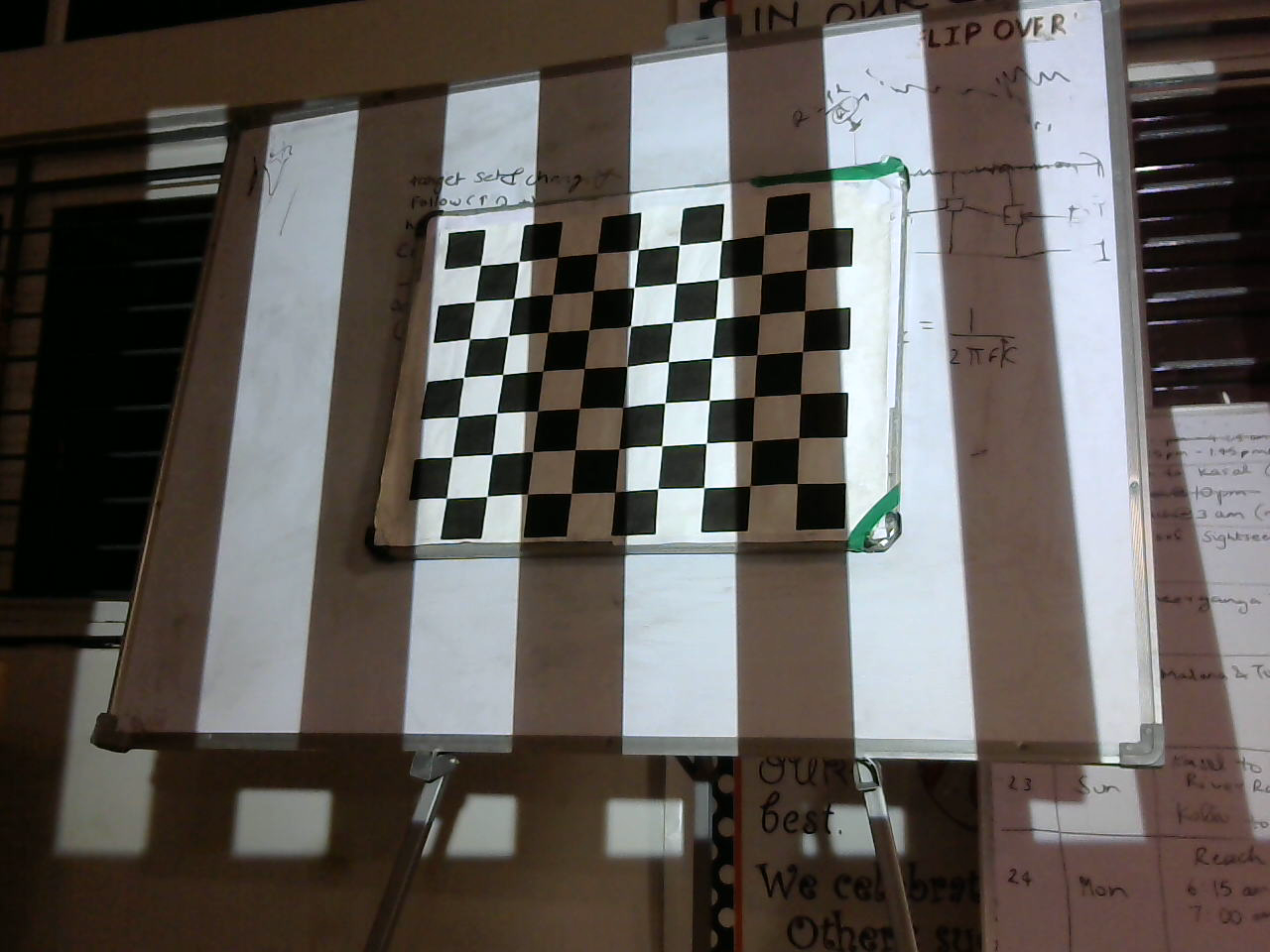}
\includegraphics[width=.4\textwidth]{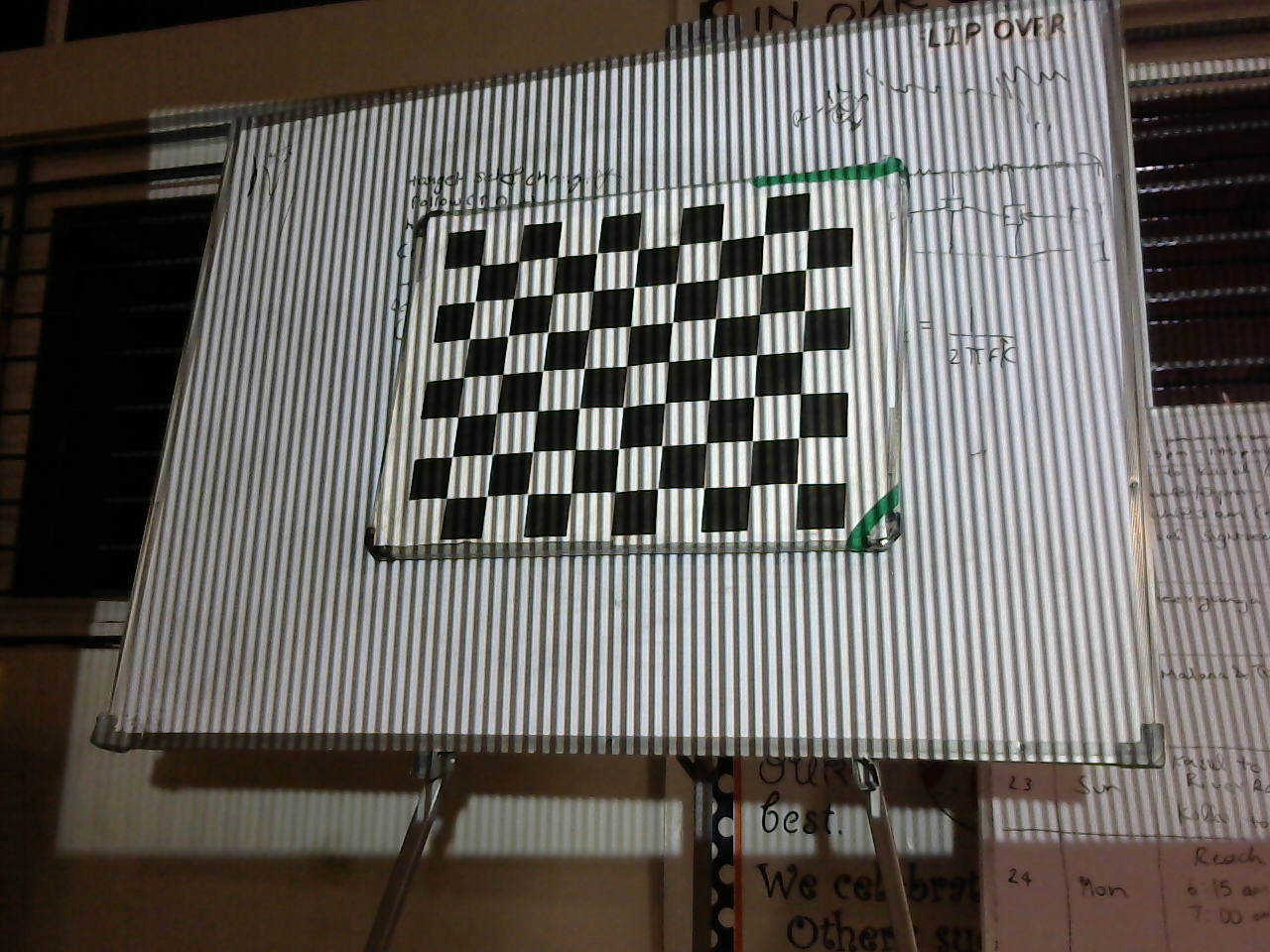}
\includegraphics[width=.4\textwidth]{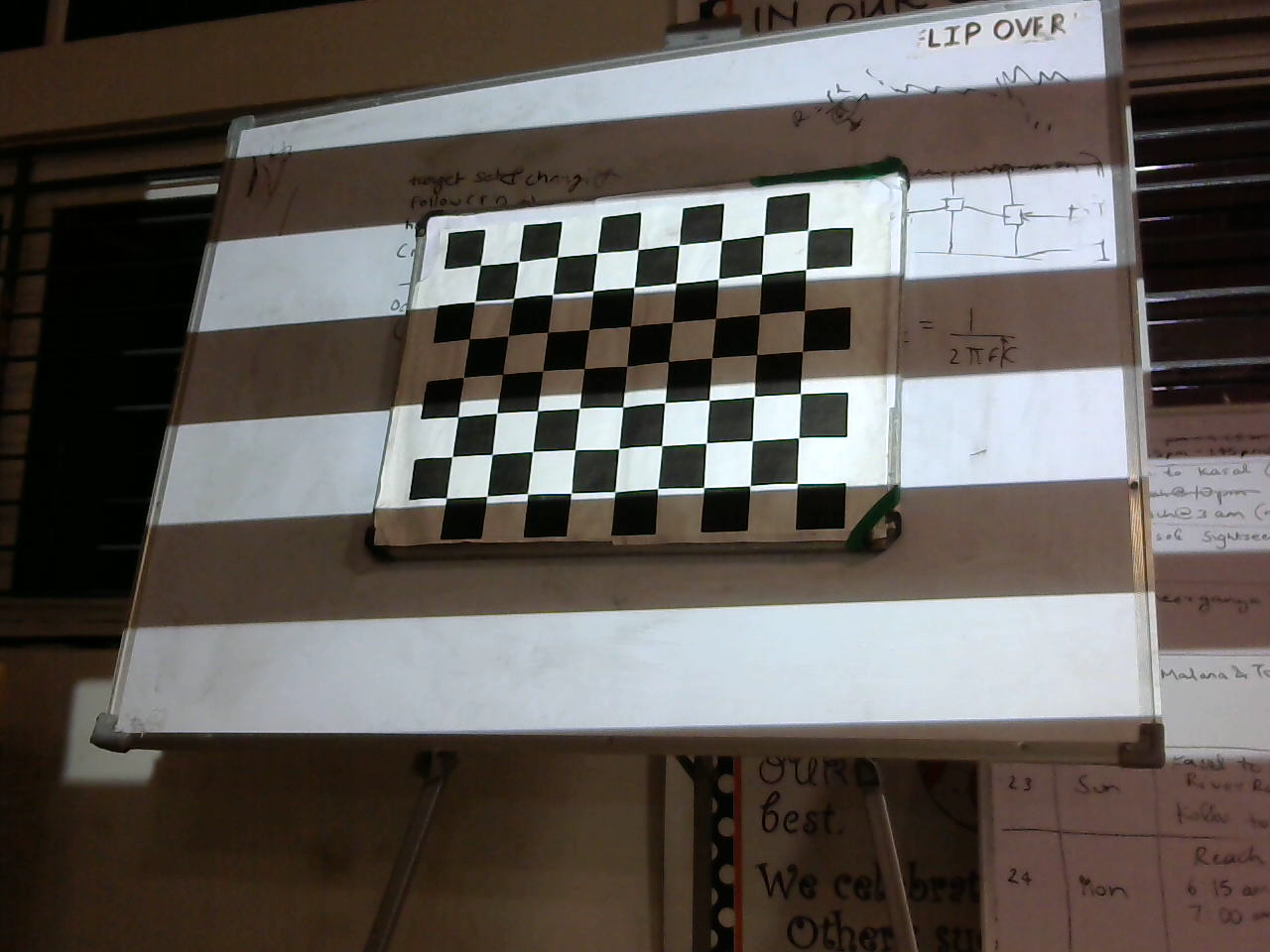}
\includegraphics[width=.4\textwidth]{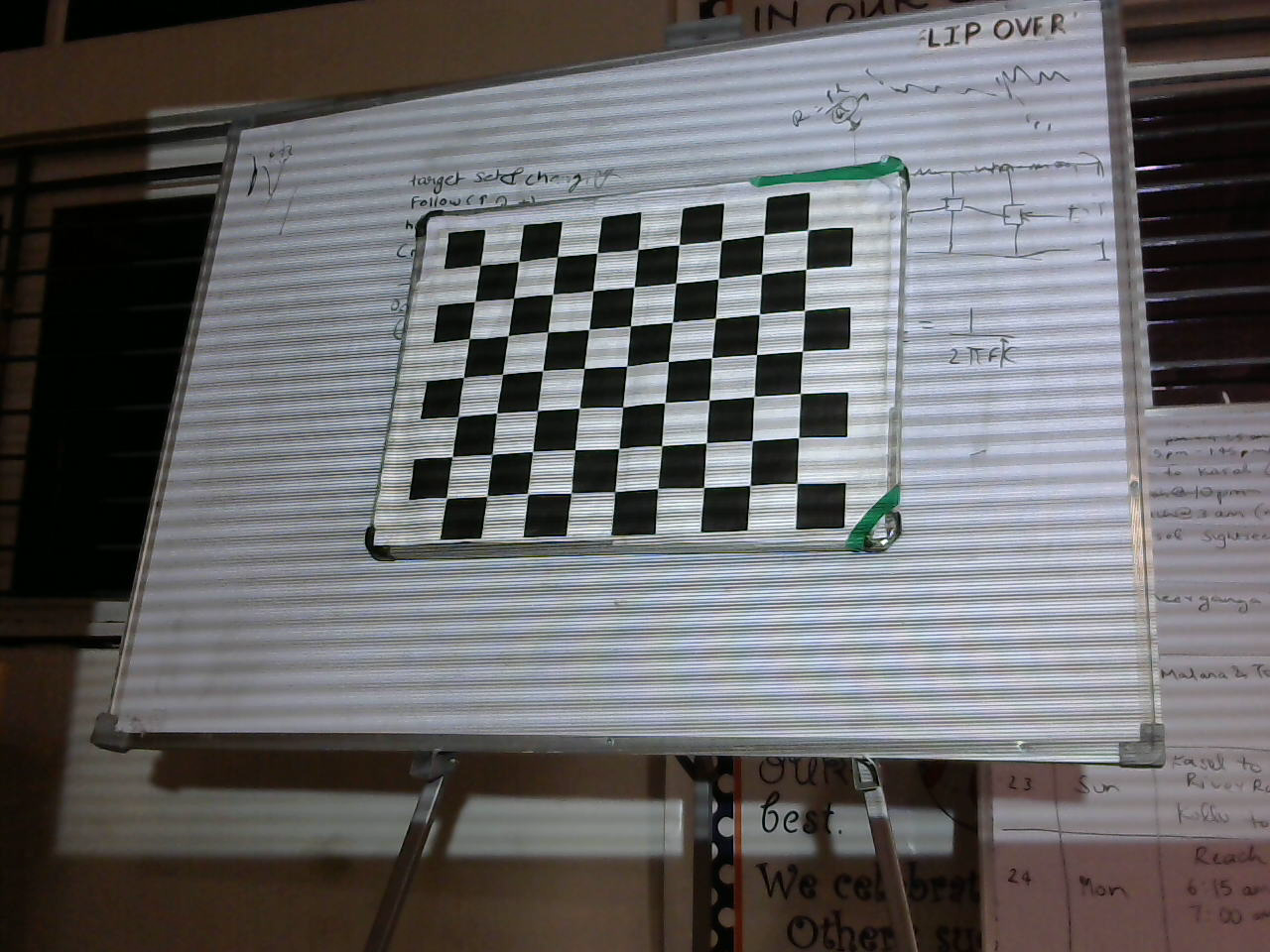}
\centering
\caption[Gray Coded Projections]{Gray Coded Projections on a real scene }
\label{fig:graycode}
\end{figure}

So far, we have only discussed how gray-coded light is encoded, we will be discussing the decoding in another chapter solely dedicated to its decoding and 3D reconstruction.

\subsection{Phase Shifted Structured Light}
In this type of structured light projection, we project horizontal and vertical sinusoidal patterns which are phase shifted by a fixed angle with respect to each other. These patterns encode the projector location in terms of phase value. The minimum number of patterns required to encode x or y-coordinate is 3, with a phase shift of 120 degrees. 
The patterns in minimal case are as follows : 
\begin{itemize}
    \item $I_1(x,y) = I'(x,y) + I''(x,y)cos(\phi - \frac{2\pi}{3})$  
    \item $I_2(x,y) = I'(x,y) + I''(x,y)cos(\phi)$  
    \item $I_3(x,y) = I'(x,y) + I''(x,y)cos(\phi + \frac{2\pi}{3})$  
\end{itemize}

Suppose, we wish to encode x-coordinate of projector pixel location using a phase value $\phi$. To do so, we will first have to decide the fringe width. Fringe width should be decided such that it is neither very large nor very small, but optimum enough that camera could adequately capture the intensity variation within a fringe. For the purpose of this setup, lets assume it to be 20 pixels. So the phase value $\phi$ would be given as $\frac{2\pi x}{20}$. 
We can decode the phase $\phi$ using the three patterns using the formula : 

$\phi(x,y) = tan^{-1}[\frac{\sqrt{3}(I_1 - I_3)}{2I_2 - I_1 - I_3}]$

The computed phase would be relative owing to the fact that range of values returned by function $tan^{-1}(x)$ will lie between $[0, 2\pi]$. So all the 'x' values decoded from phase computed using the formula $\frac{20 * \phi}{2\pi}$ would be periodic and vary from [0,20]. In order to remove this periodicity and make the values lie from [0,1920] we must add fringe number K(x,y). This would make the phase lie between [0,192$\pi$] as follows :
\\
$\Phi(x,y) = \phi(x,y) + K(x,y) * 2\pi$ , with values of K ranging from (0,96)
\\
The value of K(x,y) is computed from temporal phase unwrapping method explained by Yatong An et.al\cite{phase_un}. In this way using phase-shifted structured light we can encode the position of projector pixel location.

\begin{figure}
\includegraphics[width=.3\textwidth]{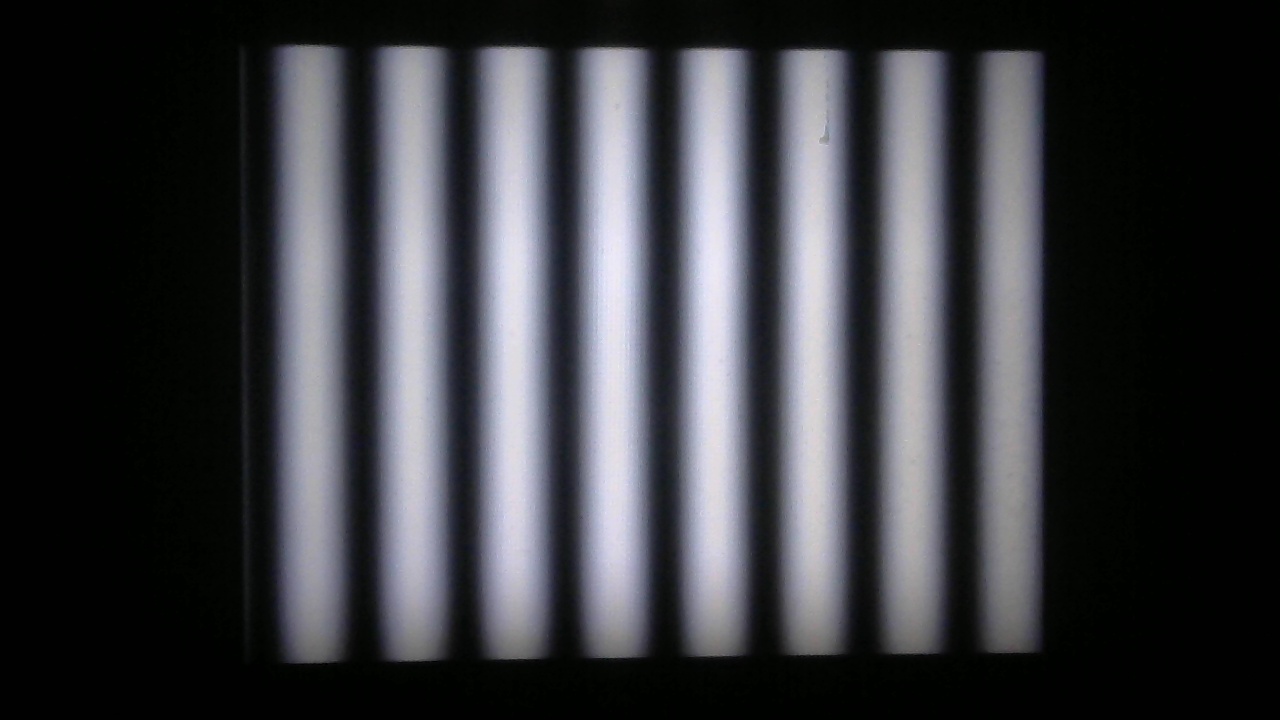}
\includegraphics[width=.3\textwidth]{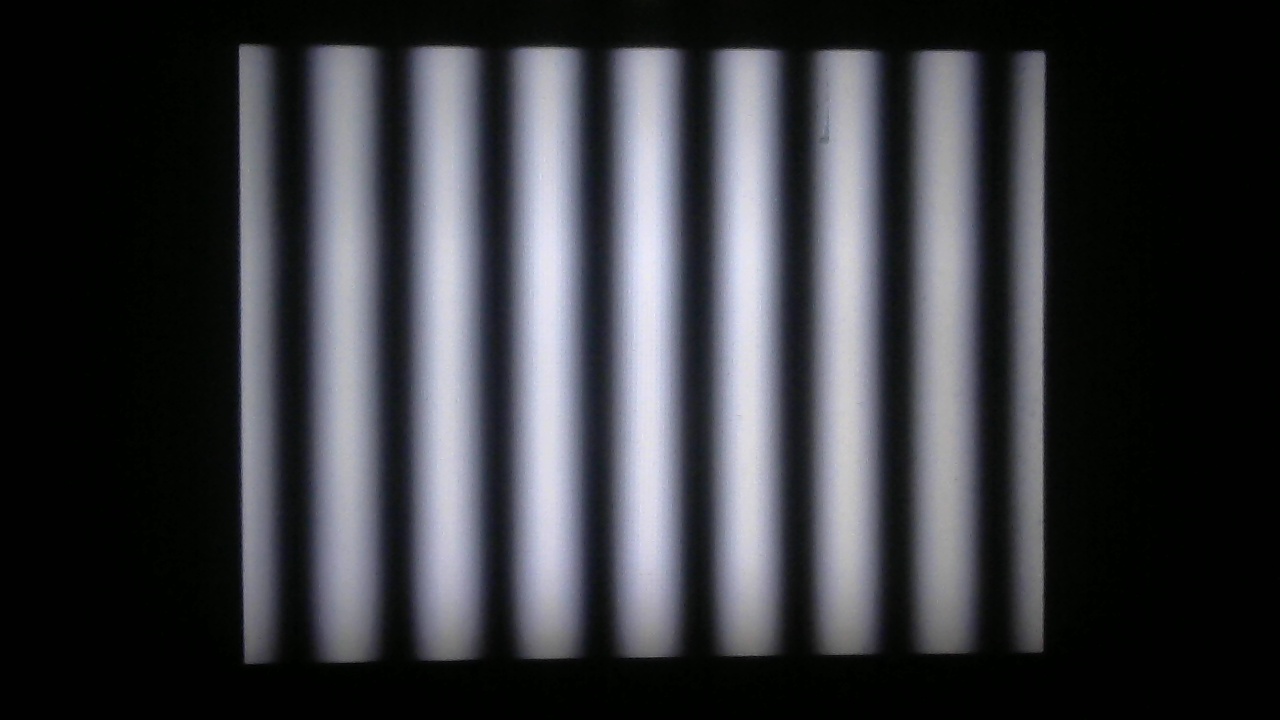}
\includegraphics[width=.3\textwidth]{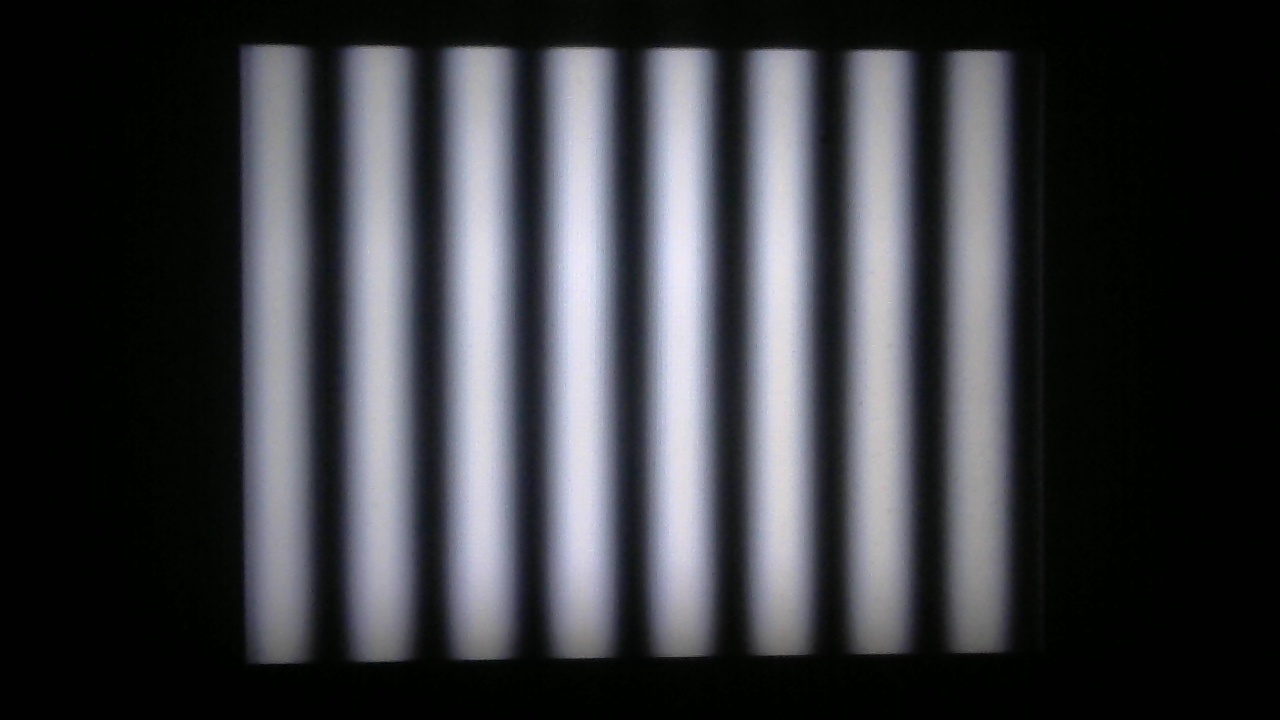}
\caption[Phase Shifted Projections]{Phase Shifted Projections on a real scene }
\label{fig:phase}

\includegraphics[width=.4\textwidth]{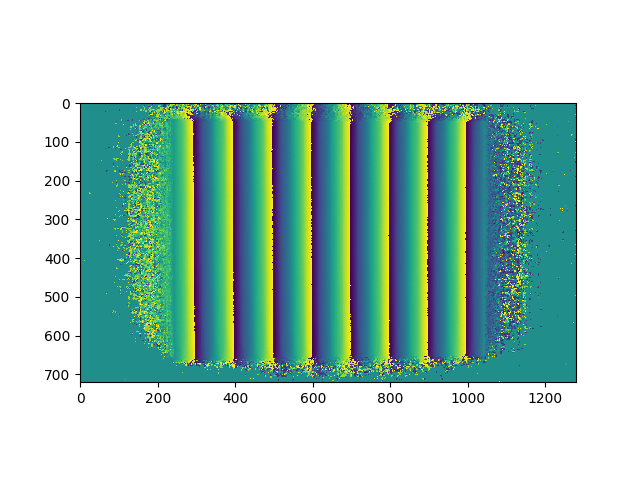}
\includegraphics[width=.4\textwidth]{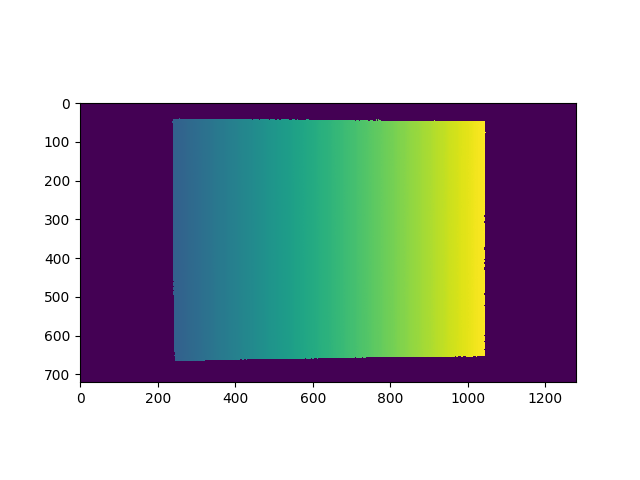}
\caption[Dewarping]{On left the image is decoded w/o phase unwrapping. On right image is decoded with phase unwrapping }
\label{fig:dewarping}

\end{figure}

The projections as well as results of decoding with and without phase unwarping are shown in the Fig \ref{fig:phase} and Fig \ref{fig:dewarping}

\subsection{Hybrid Structured Light} In the hybrid method we take advantage of both gray-coded structured light and phase-shifted structured light. Decoding in gray-coded algorithm do not work well as the frequency increases. Also, there are issues of direct and indirect light, inter-reflections, etc in the scene. This could be handled from phase-shift as it gives us the estimate of indirect light in the scene. Also phase-shift algorithm can give us correspondences with sub-pixel accuracy. However, there are issues with phase unwrapping in phase-shift method, also it do not work well with lower frequencies. So, we can generate decoding from gray-code while working on lower frequencies and use them to generate K(x,y) and then operate on higher frequencies with phase-shifted light. However, due to unforeseen circumstances, we were not able to implement this algorithm.

\chapter{Projector and System Calibration}
\label{chapter:projector}

\section{Overview}
\par The structured light method is a very simple yet effective method of gathering 3D models. They require easily available and cheap off-the-shelf components; a projector and a camera and are still able to produce comparable results with that of expensive laser scanners. Accurate results however require accurate calibration of both projector and camera. There are an assortment of robust camera calibration methods available and both projector and camera can be described using the same model. However, camera calibration methods cannot be well adapted to projector calibration and therefore a simplified model is used which is not very precise.

The method applied here uses an uncalibrated camera and local homographies to achieve sub-pixel precision. This way the projector can be described using any camera model which includes the extended pinhole model.

\section{Introduction}

A system based on structured light is the best choice for a do-it-yourself application. This is because they are easily deployable and only require the use of a projector and a camera. The projector and camera form and work as a stereo system, and if the projected pattern is carefully chosen then the job of finding point correspondences is simplified. The projector in such a system is modeled as a reverse camera and all the passive stereo system considerations are applied almost as it is.
The projectors however cannot measure the pixel coordinates of 3D points directly, which are projected onto the projector image similar to cameras, and thus the calibration procedure must be modified. Parameters of both projector and camera such as viewpoint, zoom, and focus must be attuned to equal the object size and scanning distance of the target object. This would nullify any previous calibration. Thus, they structured light system requires calibration before every use for it to guarantee optimum results. This mandates the simplicity of the calibration procedure along with its precision.  Presented here is a easy to perform and very accurate method of structured light calibration.
The estimation of the coordinates of calibration points using local homographies in the projector image plane is the chief impression of this method\cite{6375029}. The first step is to find a dense set of correspondences between projector and camera pixels by projecting an identical pattern sequence as the one later projected to scan the target onto the calibration object. This reuses a majority of software components that were written for the scanning application. Next, a group of local homographies are computed using the set of correspondences which allow to find with sub-pixel precision the projection of any of the points in the calibration object onto the projector image plane. Finally, the calibration of data projector is done as a normal camera.
The central feature is the method of obtaining correspondences between projector pixels and 3D world points. Any calibration technique which is offered for passive stereo can be used directly to structured light system after the correspondences are known.
In consequence the accuracy of camera calibration does not affect the calibration of the projector in any way. This calibration technique does not require user intervention once data acquisition is completed which renders its operation possible even by users without much experience.

\section{Literature Review}
There previously exist various other calibration procedures, however none of them are easy to use and lack adequate precision to facilitate accurate 3D reconstruction. An assortment of methods involve a camera that is calibrated beforehand which is then used to allocate projector correspondences.
Such methods are easy to perform, however they do not display accuracy in projector parameters due to their heavy dependency on camera calibration. The inaccuracies occur due to the approach used, since even small calibration errors cause large world coordinate errors. They fail because they estimate projector parameters from world coordinates which are not very accurate, leading to a decrease in precision of the entire system.
Some methods adopt a different approach where a calibrated camera and printed pattern is not required. They ask the user to move the projector across several positions so that the projected pattern which is projected onto a fixed plane, changes its shape. It is however, inconvenient or impossible to move the projector in general and such a system cannot be used if metric reconstruction is required.
Other methods involve a projected pattern which is iteratively adjusted to overlap a printed pattern. An uncalibrated camera is used to measure the overlap.  This requires clear identification of both patterns, thus colour patterns are used instead of black and white and also a colour camera is required. Using colour patterns mandated colour calibration but the printed and camera colours rarely match. This technique also requires continuous input from a camera which makes it impossible to isolate the capture stage from calibration algorithm.
Other calibration methods commonly involve homography transformation between calibration plane and the plane of the projector image. But since they are linear operators they cannot model the non-linear lens distortions that are introduced by projector lenses. One method that involves projecting patterns on a flat aluminium board mounted on a high precision moving mechanism gets very accurate results but such special equipment is not available to the common user.
Zhang and others use structured light and instead of getting projector point correspondences from camera images, they make new synthetic images from the viewpoint of the projector which are then fed to standard camera calibration tools. This step of creating synthetic images might lead to ignorance of some vital information which is detrimental. The method used here get projector point correspondences from structured light patterns at the camera resolution.

\section{Proposed Method}
The setup consists of a projector and a camera which behave as a stereo pair. They are both described using the pinhole model extended with radial and tangential distortion. This gives an advantage over various other methods which do not account for distortion of projected patterns, as most projectors have distortions present outside their focus plane which affect the accuracy of the final 3D models.

Zhang’s method was followed closely since it boasts simplicity along with great accuracy. It makes use of a plane checkerboard for calibration, which can easily be printed onto paper. The method follows the user capturing images of the checkerboard in various positions or orientations and then the algorithm is used to calculate camera calibration parameters by making use of relation between world coordinate system which is attached to the checkerboard plane and the checkerboard corners in a camera coordinate system.

Since the projector and the camera use similar models we have used the same calibration procedure for the projector as was used for the camera.

However there is a difference between the two as the projector cannot capture images like the camera, of the checkerboard pattern, hence a slightly different technique has to be used.
We have already extracted a relation between projector and camera pixels using the structured light approach, using this we can estimate the checkerboard corner locations in projector pixel coordinates.

(The act of estimating intrinsic or extrinsic parameters of a camera is termed as camera calibration. The intrinsic values of a camera include its focal length, distortion, skew et cetera. The extrinsic values are those which are used to describe the position and orientation of the camera in the world. Zhang’s method is to be used according to which the coordinates of every checkboard corner are to be found in the image plane. A standard procedure is used to locate the corner positions of every checkerboard orientation.)

The computation of the coordinates of the checkerboard corners through the projector coordinate system is done in three steps:
Initially the decoding of structured light sequence is first carried out and each pixel of camera is linked with a row and column of projector. Next estimation of homography of every checkerboard corner in camera image is carried out. Finally, through application of obtained homography every corner of camera coordinate is transformed to projector coordinate.
 
The decoding of structured light is hinged on the projected pattern which is the grey code for rows and columns. The method used here favours decoding precision over acquisition speed. A part of grey code images where the stripes look narrow can be regarded to exhibit a high frequency pattern. These enable the splitting of the intensity measured at every pixel in a direct and global component\cite{article}.
 
The light received at each camera pixel is to be affected by exactly one pixel of a projector, but this only occurs only ideally. But in operation a camera pixel is affected by the total of light from projector pixel along with light from various other sources. An inability to correctly categorize or label these sources or components of light leads to decoding errors. The correct identification of each component if done then a simple set of rules can be used for a radical reduction in decoding errors\cite{Xu2007RobustPC}.
The use of structured light relation to convert camera coordinates to projector coordinated is still not possible due to the absence of their bijective nature. Local homography can be used to counter this issue which is applicable only in a part of the plane. Thus, a local homography for every corner of the checkerboard is found instead of a solitary global one.
The estimation of every local homography is done within the area of a selected corner and can only be used to convert that particular corner to projector coordinates.
The modelling of non-linear distortions is possible due to the individual translation of every corner. Also they lead to a reduction of small decoding errors since their estimation is derived through more points than are required.

\chapter{Triangulation}
\label{chapter:Triangulation}

\section{Overview}
Triangulation is the process of reconstructing 3D points from multiple 2D image projections. To do so,it is essential that camera and projector are calibrated, i.e their intrinsic matrices are known, as well as their relative rigid-body transformation, commonly known as extrinsic relation is also known. Once, all this information is available it is very easy to perform triangulation\cite{hartley1997triangulation}. 

\section{Linear Triangulation}
Once, we finish decoding our Structured Light pattern, we will get correspondences between camera and a projector. Also, we have already calibrated our camera intrinsics $K_{camera}$, projector intrinsics $K_{projector}$ and extrinsic registration between camera and projector composed of Rotation matrix R and translation vector t. So now we can generate two projection matrices as follows : 
\begin{itemize}
    \item Assume either projector's Center of Projection or Camera's Center of Projection to be origin of world co-ordinate system. We will assume camera's COP as our origin.
    \item Since we know the relative position of camera and projector, we can write the rigid body transformation between camera and world as well as between projector and world as follows :
    \\
    \\
    $ X^{camera} = \begin{bmatrix} 1 & 0 & 0 \\ 0 & 1 & 0 \\ 0 & 0 & 1 \end{bmatrix} X^{world} + \begin{bmatrix} 0\\ 0\\ 0 \end{bmatrix}$;
    \enspace
    $X^{projector} = \begin{bmatrix} r_{11} & r_{12} & r_{13} \\ r_{21} & r_{22} & r_{23} \\ r_{31} & r_{32} & r_{33} \end{bmatrix} X^{world} + \begin{bmatrix} t_x\\ t_y\\ t_z \end{bmatrix}
    $ 
    
    \item The relationship between the image-coordinates and world coordinates for camera and projector can now be written as using pin-hole model described in Chapter \ref{chapter:Camera_Calibration} as follows: 
    \begin{equation}
    \lambda_{camera} \begin{bmatrix}u_c \\ v_c \\ 1\end{bmatrix} = K_{camera}\begin{bmatrix} 1 & 0 & 0 & 0 \\ 0 & 1 & 0 & 0\\ 0 & 0 & 1 & 0\end{bmatrix} \begin{bmatrix} X^{world}\\ 1 \end{bmatrix}
    \end{equation}
    
    \begin{equation}
    \lambda_{projector} \begin{bmatrix}u_p \\ v_p \\ 1\end{bmatrix} = K_{projector}\begin{bmatrix} r_{11} & r_{12} & r_{13} & t_x \\ r_{21} & r_{22} & r_{23} & t_y\\ r_{31} & r_{32} & r_{33} & t_z \end{bmatrix} \begin{bmatrix} X^{world}\\ 1 \end{bmatrix}
    \end{equation}
    \\
    Here projection matrix for the camera would be denoted by $P_{camera}$ and would be equal to :
    $P_{camera} = K_{camera}\begin{bmatrix} 1 & 0 & 0 & 0 \\ 0 & 1 & 0 & 0\\ 0 & 0 & 1 & 0\end{bmatrix}$ \\
    and similarly, projection matrix for the projector would be denoted by $P_{projector}$ and will be equal to  $P_{projector} = K_{projector}\begin{bmatrix} r_{11} & r_{12} & r_{13} & t_x \\ r_{21} & r_{22} & r_{23} & t_y\\ r_{31} & r_{32} & r_{33} & t_z \end{bmatrix}$. 
    \\
    Let the correspondence in camera and projector be denoted by $x_{camera} =  \begin{bmatrix}u_c & v_c\end{bmatrix}^T$ and $x_{projector}=  \begin{bmatrix}u_p & v_p \end{bmatrix}^T$ respectively. So now we can re-write equation 4.1.1 and 4.1.2 as 
    \begin{equation}
        \lambda_{camera} \begin{bmatrix}x_{camera} \\ 1\end{bmatrix} = P_{camera}\begin{bmatrix}X_{world} \\1\end{bmatrix}
    \end{equation}
    \begin{equation}
        \lambda_{projector} \begin{bmatrix}x_{projector} \\ 1\end{bmatrix} = P_{projector}\begin{bmatrix}X_{world} \\1\end{bmatrix}
    \end{equation}
\item Since left-side of equation 4.1.3 and right-side of equation 4.1.3 points in same direction. Their cross-product must be zero. Therefore, \\
\begin{equation}
\lambda_{camera} \begin{bmatrix}x_{camera} \\ 1\end{bmatrix} \times P_{camera}\begin{bmatrix}X_{world} \\1\end{bmatrix}   = 0  
\end{equation}
Similarly, from equation 4.1.4, we can write\\
\begin{equation}
\lambda_{projector} \begin{bmatrix}x_{projector} \\ 1\end{bmatrix} \times P_{projector}\begin{bmatrix}X_{projector} \\1\end{bmatrix}   = 0  
\end{equation}

On opening the equation 4.1.5 we get,
\begin{equation}
    \begin{bmatrix} u_c \\ v_c \\ 1\end{bmatrix} \times \begin{bmatrix} p_1^T X_{world} \\ p_2^T X_{world}\\p_3^T X_{world}\end{bmatrix} = \begin{bmatrix} v_c p_3^T X_{world} - p_2^T X_{world}\\ p_1^T X_{world} - u_c p_3^T X_{world}\\ u_c p_2^T X_{world} - v_c p_1^T X_{world} \end{bmatrix} = 0
\end{equation}
The third row is linearly dependent on first and second row, therefore, we get two equations from one correspondence. Similarly on opening equation 4.1.6 for the corresponding point in projector we will get two more equations. Thus for three unknowns in $X_{world}$ we have 4 equations. We find the least-square solution for all the four equations. To do so,  let us denote $P_{projector}$ as $\begin{bmatrix} q_1^T & q_2^T & q_3^T\end{bmatrix}^T$. So the 4 equations will be:
\begin{equation}
    \begin{bmatrix} v_c p_3^T  - p_2^T \\ p_1^T  - u_c p_3^T \\ v_p q_3^T  - q_2^T \\ q_1^T  - u_p p_3^T  \end{bmatrix} X_{world} = 0
\end{equation}

The least square solution can be obtained by performing the Singular Value Decomposition(SVD) of matrix mentioned in equation 4.1.8. The right singular vector corresponding to smallest singular value is the valid 3D point corresponding to the correspondences in camera and projector.

\begin{figure}[h]
\includegraphics[width=\textwidth]{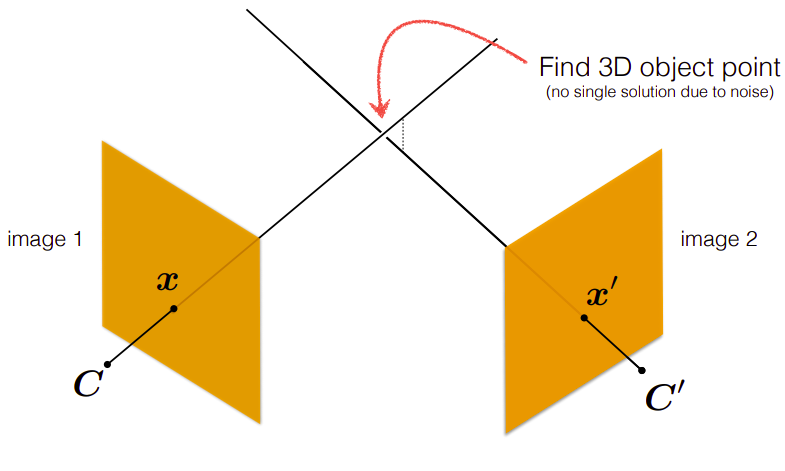}
\centering
\caption[Linear Triangulation]{Linear triangulation with least square error }
\label{fig:triangulate}
\end{figure}
The whole process can be explained with the help of a Figure \ref{fig:triangulate}. Here C represents the COP of camera and C' represents the COP of Projector. x represent the projection of some 3D point in camera-image plane and x' represent the projection of the same 3D point in projector-image plane. However, due to noise, the correspondences may be wrong and the rays may not intersect as shown in the figure. Therefore, multiple solutions are possible, in order to resolve this problem, we try to find the point that lies on the midpoint of the shortest distance between the two rays. This is mathematically done using least-squares method.
\end{itemize}

\subsection{Non-Linear Refinement}
There are strategies which performs non-linear triangulation given camera-projector correspondences. These methods takes the result from linear-triangulation as the initial estimate and performs non-linear refinement. They try to minimize reprojection error in image as well as in 3D space. We wont be going into details of this, since we didn't get a chance to implement this algorithm. However, it would have certainly improved the 3D point cloud.

\chapter{Point Cloud Registration}
\label{chapter:icp}

\section{Overview}
Since we are using turntable like setup, for every angle of rotation on turn table we will get one point cloud. We in our experiment have used 36 point clouds captured in intervals of 10 degrees. Now in order to stitch all the point clouds we will need to find the rigid body transformation linking the two point clouds. This can be elegantly done using Iterative Closest Point(ICP) algorithm \cite{chetverikov2002trimmed}. With this algorithm we would be able to stitch all the point clouds together incrementally.

\section{Estimation of Rigid Body Transformation with known Correspondences}

Given two point clouds with overlapping scene structures, a source \textbf{P} and a target \textbf{X}, the ICP algorithm estimates the rigid transformation matrix that aligns/registers a given pair of point clouds (target \& source). This transformation could be used as a piece of odometry information to accomplish localization and planning. Also with the help of these transformations, we can create a large scale 3D map by stitching multiple point clouds. This could be mathematically modeled as follows:

Initially, let us assume that there are two point clouds with N known correspondences, (we generally always don't have such correspondences) given by :
\\
$X = {x_1, x_2, x_3, ......., x_N}$ and $X = {p_1, p_2, p_3, ......., p_N}$. Here X and P are $3 \times N$ matrices
Our aim is to find the rotation matrix \textbf{R} and a translation vector \textbf{t} that will minimize :
\begin{equation}
e = \frac{1}{N}\sum_{i=1}^N ||x_i - Rp_i - t||^2   
\end{equation}

The first step is to subtract the mean from each point cloud so that the mean of the new point clouds is zero. This can be done as:
\begin{equation}
    x_i' = x_i - \mu_x \enspace \text{where} \enspace \mu_x = \frac{1}{n}\sum_{i=1}^N x_i
\end{equation}
\begin{equation}
    p_i' = p_i - \mu_p \enspace \text{where} \enspace \mu_p = \frac{1}{n}\sum_{i=1}^N p_i
\end{equation}
On substituting this new point clouds in error term we get,
\begin{equation}
e = \frac{1}{N}\sum_{i=1}^N ||x_i' - Rp_i' - t'||^2 \enspace where \enspace t' = t - \mu_x - R\mu_p   
\end{equation}
On opening the error term, we get:
\begin{equation}
e = \frac{1}{N}\sum_{i=1}^N (x_i' - Rp_i' - t')^T(x_i' - Rp_i' - t') 
\end{equation}

\begin{equation}
e = \frac{1}{N}\sum_{i=1}^N (x_i'^Tx_i' - x_i'^T(Rp_i') - x_i'^Tt' - (Rp_i')^Tx_i' + (Rp_i')^T(Rp_i') + (Rp_i')^Tt' -t'^Tx_i' + t'^T(Rp_i') + t'^Tt')
\end{equation}

On applying the summation and using the property that the mean of  X' and  P' is zero and t' is constant, we will get:
\begin{equation}
e = \frac{1}{N}\sum_{i=1}^N (||x_i'||^2 + ||p_i'||^2 + ||t'||^2 - 2x_i'^TRp_i')    
\end{equation} 
From the above equation, it is evident that in order to have the minimum error we have to minimize the norm of t' and maximize the last term.
The minimum norm of translation t' can be at max zero. So the optimal translation that will minimize the error term would be given as  :
\begin{equation}
||t'||^2 = 0 \enspace thus,  \enspace t' = 0 \enspace \therefore t = \mu_x - R\mu_p
\end{equation}
Now, if we know the rotation matrix R we can easily find the translation vector t.

To find rotation matrix R we will try to maximize the last term. The last term can be rewritten as 
\begin{equation}
\frac{1}{N}\sum_{i=1}^N 2x_i'^TRp_i' = 2tr(X (RP)^T)
\end{equation}
\begin{equation}
tr(X (RP)^T) = tr(X P^T R^T)
\end{equation}
Let,
\begin{equation}
Q = XP^T \enspace and \enspace R' = R^T
\end{equation}
\begin{equation}
    SVD(Q) = U S V^T
\end{equation}
\begin{equation}
    \frac{1}{N}\sum_{i=1}^N 2x_i'^TRp_i = tr(U S V^T R') = tr(V^T R'U S) 
\end{equation}
Let, 
\begin{equation}
    W = V^T R'U
\end{equation}
Here, W is an orthogonal matrix since it is a composition of other orthogonal matrices. W being an orthogonal matrix the maximum entry in the W matrix will be 1. However, in order to maximize this term, all the diagonal entries should be 1. Therefore, the W matrix that will maximize this term would be an identity matrix. Thus, Rotation matrix R can be estimated as follows :
\begin{equation}
    W = V^T R'U = I  
\end{equation}
\begin{equation}
    R' = VU^T  
\end{equation}

\begin{equation}
    R = UV^T  
\end{equation}
Thus, R can be estimated from the SVD of matrix Q, which is already known from point clouds X and P. After estimating R we can also compute translation vector t from the mean of X and P. In this way with known correspondences we can estimate the rigid body transformation R and t.

\section{Estimation of unknown Correspondences}

Now that we have established a mathematical model to estimate the rigid body transformation with known correspondence, we will build on this model to solve the problem if correspondences are not known.

In the case of unknown correspondences, we first estimate the best matches between the two point clouds using the following methods :
\begin{itemize}
    \item closest points (in terms of euclidean distance)
    \item normal shooting
    \item projective data association
\end{itemize}   

There are other methods too, but the aforementioned methods are most commonly used.

\subsection{Closest Point Method} 

In this method,  we initially assume zero translation and identity rotation between the two point clouds. Now,  for every point in one point-cloud, we find the closest point in the other point cloud.  This is usually done using k-d trees and is slower.

\begin{figure}
    \centering
    \includegraphics[width=\textwidth]{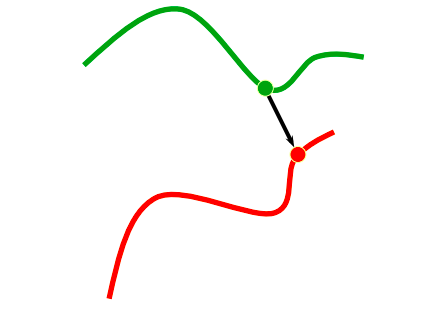}
    \caption{Closest Point Method}
    \label{fig:closest_point}
\end{figure}

\subsection{Normal Shooting}
In this method, for every point in one point cloud, a normal is estimated and the closest point in the direction of normal in the other point cloud is considered as correspondence. It is faster than closest point method.

\begin{figure}
    \centering
    \includegraphics[width=\textwidth]{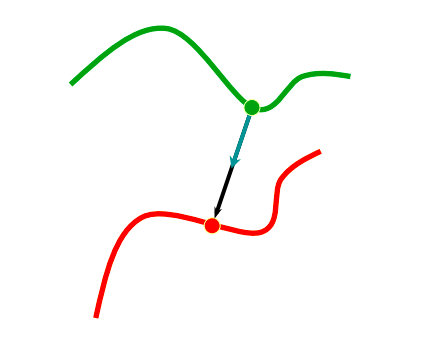}
    \caption{Normal Shooting}
    \label{fig:normal_shooting}
\end{figure}

\subsection{Projective Data Association}
This method is commonly used in RGBD cameras, LIDAR, etc which have known ray directions. It is the fastest method since no computation is required. In this method, the closest point is in the direction of the ray emitted from the camera/LIDAR. We are using this method.
\begin{figure}
    \centering
    \includegraphics[width=\textwidth]{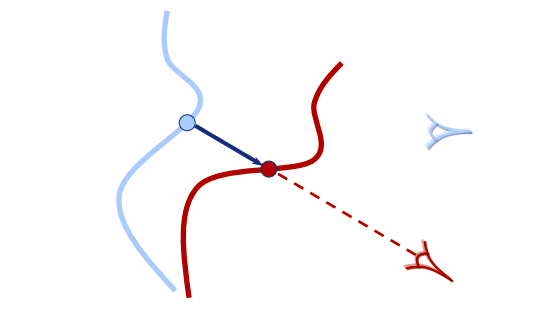}
    \caption{Projective Data Association}
    \label{fig:projective}
\end{figure}

\section{Error Metrics}

ICP as the name suggests is an iterative algorithm. In order to estimate how well the algorithm is doing and its termination criteria, it is essential to have an error metric. The two most commonly used error metrics are :
\begin{itemize}
\item \textbf{Point-Point error} - The sum of the euclidean distance between all the corresponding points in the two point clouds is known as Point-Point error. This error metric causes slower convergence. It is the same error that we have used in the previous section to estimate R and t.
\item \textbf{Point-Plane error} -  It is defined as the sum of the squared distance between all the source points and the tangent planes in their corresponding destination point-cloud.
\begin{equation}
    e = \frac{1}{N}\sum_{i=1}^{N}((x_i - Rp_i - t).n_i)^2
\end{equation}
Here, $n_i$ is the unit normal vector at that point.
\end{itemize}

\textbf{Note: Since we have found the optimal translation and rotation for point-point error metric we will be using this metric in all the further steps of the algorithm. However, if you wish to use a point-plane error metric you will have to recompute the expression for optimized rotation and translation that will minimize that metric.}

\section{Algorithm}
Now that we are versed with the mathematical background, we can perform the steps of the ICP algorithm defined below :
\begin{itemize}
    
\item Assuming zero translation and identity rotation, compute the correspondences using any of the methods mentioned above depending on the complexity of the problem you are trying to solve.
\item Using those correspondences estimate the rigid body transformation R and t that will minimize the point-point error metric
\item If the error is below the pre-defined threshold, meaning you have achieved sufficient accuracy, terminate the algorithm, else, with the new estimated transform, repeat steps 1 to 3.

\end{itemize}

\section{Limitations}
Algorithms like ICP are only used when the rigid body transformation between two point clouds is small or the transformation is known. Also, if the correspondences are not estimated correctly, the algorithm might not converge. 

\chapter{Results and Conclusion}
\label{chapter:Results}

In this chapter results after every steps are presented.

\section{Results after camera calibration}

$K_{camera}$ = \begin{table}[hbt!]
\begin{tabular}{ | c | c| c | } 
\hline
1.44038101e+03 & 0.00000000e+00 & 6.67836875e+02 \\ 
\hline
0.00000000e+00 & 1.43711605e+03 & 3.54202552e+02 \\
\hline
0.00000000e+00 & 0.00000000e+00 & 1.00000000e+00
\\ 
\hline
\end{tabular}
\caption[Intrinsic Matrix of Camera]{Intrinsic Matrix of Camera}

\end{table}

$D_{camera}$ = \begin{table}[hbt!]
\begin{tabular}{ | c | c| c | c | c|} 
\hline
5.4658e-01 & -2.0200e+01 & -2.2032e-02 & 8.2588e-03
  & 2.1111e+02\\ 
\hline
\end{tabular}
\caption[Distortion coefficients of Camera]{Distortion coefficients of Camera}

\end{table}

To verfiy the accuracy of camera calibration, we estimate its reprojection error. It should be less than 1 to ensure valid calibration.

\textbf{Reprojection error = 0.5997 pixels}

\section{Results after projector calibration}
Since projector is calibrated using the same model as camera, similar matrices for projector are also computed.

$K_{projector}$ = \begin{table}[H]
\begin{tabular}{ | c | c| c | } 
\hline
4.61896471e+03 & 0.00000000e+00 & 7.44248184e+02 \\ 
\hline
0.00000000e+00 & 4.34710132e+03 & 6.05056615e+02 \\
\hline
0.00000000e+00 & 0.00000000e+00 & 1.00000000e+00
\\ 
\hline
\end{tabular}

\caption{Intrinsic Matrix of Projector}
\end{table}

$D_{projector}$ = \begin{table}[htb!]
\begin{tabular}{ | c | c| c | c | c|} 
\hline
-6.41035436e-01 & 1.24352939e+01 & -6.08400958e-02 & -3.44770205e-03 & 2.01572896e+02\\ 
\hline
\end{tabular}
\caption{Distortion coefficients of Projector}

\end{table}

To verfiy the accuracy of projector calibration, we estimate its reprojection error. It should be less than 1 to ensure valid calibration.

\textbf{Reprojection error = 0.7117 pixels} (< 1.0)

\section{Extrinsic Registration}
With respect to some arbitrary world cordinate system the extrinsic parameters of camera and world are as follows:

$T_{camera}$ = \begin{table}[hbt!]
\begin{tabular}{ | c | c| c | c| } 
\hline
0.96523178 & -0.06801648 &  0.25239131 & -257.85891969 \\
\hline
-0.05246011 & 0.89550287 & 0.4419531 & -53.58173962\\
\hline
-0.25607724 &  -0.43982765 & 0.86079968 & 1944.10591591 \\
\hline
\end{tabular}
\caption{Transformation matrix between camera and arbitrary world co-ordinate system}

\end{table}

$T_{projector}$\begin{table}[hbt!]

\begin{tabular}{ | c | c| c | c |} 
\hline
0.94278236 & -0.06015904 & 0.32793645 & -120.20727633\\
\hline
-0.03480907 & 0.96045107 & 0.27626448 & 335.26497471\\
\hline
-0.33158673 & -0.27187244 & 0.90340225 & 4481.52560352 \\
\hline
\end{tabular}
\caption{Transformation matrix between projector and arbitrary world co-ordinate system}

\end{table}

\section{Decoding Robust Gray-code}
\begin{figure}[hbt!]
    \centering
    \includegraphics[width=\textwidth]{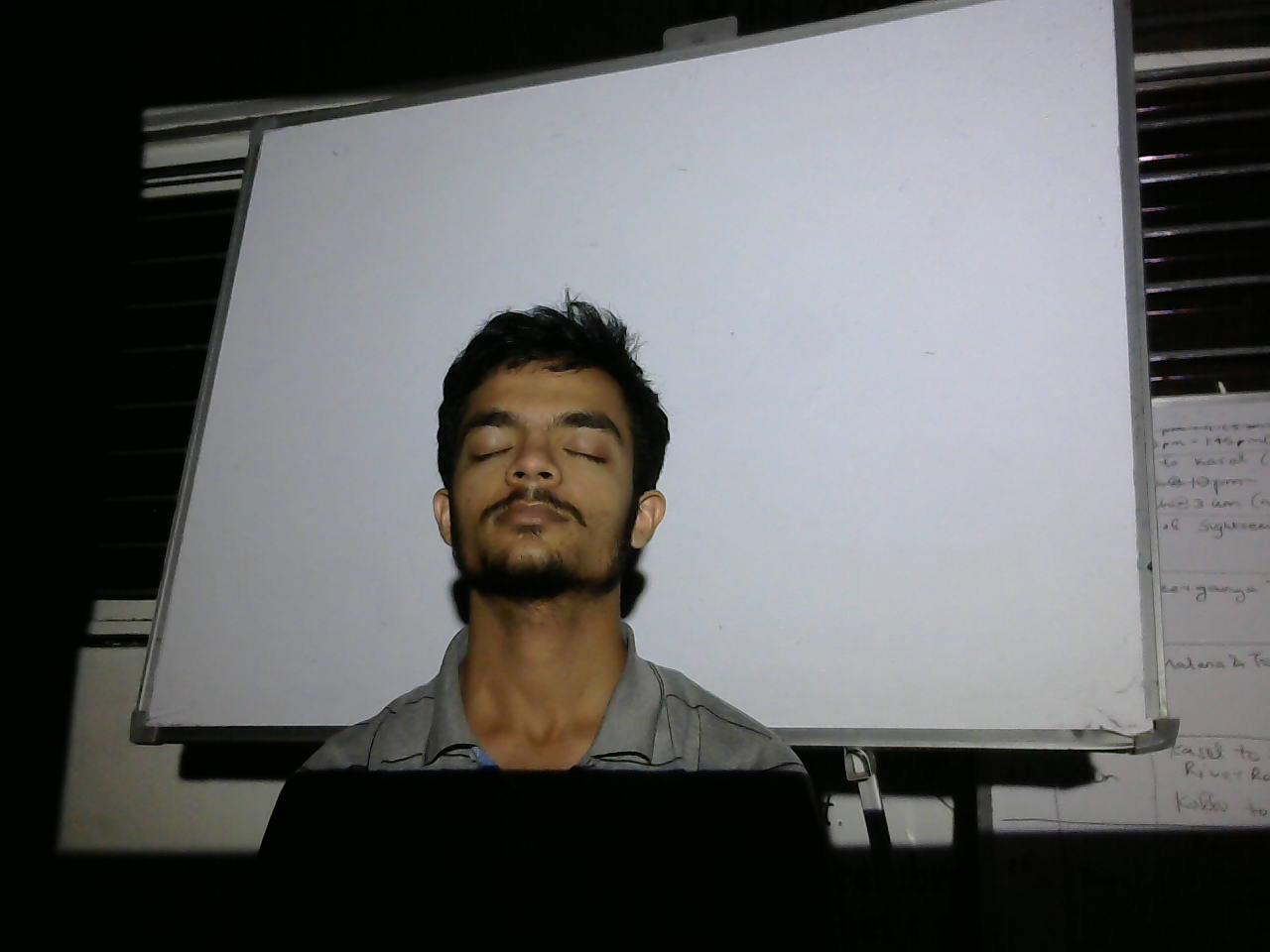}
    \caption{Scene Image}
    \label{fig:scene}
\end{figure}

\begin{figure}[hbt!]
    \centering
    \includegraphics[width=\textwidth]{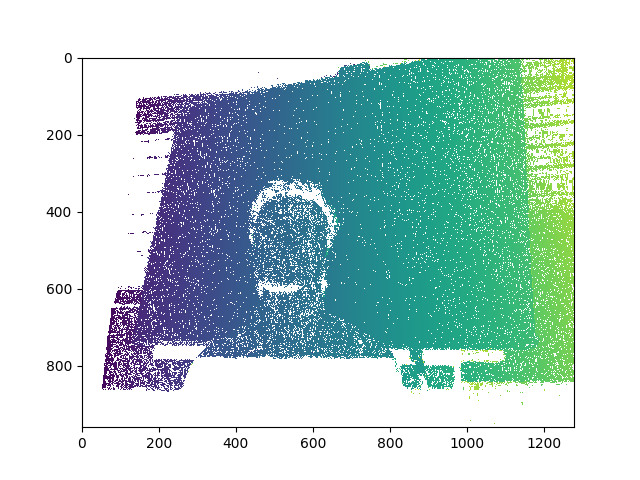}
    \caption{Decoding in x-direction using graycoded SL}
    \label{fig:x}
\end{figure}

\begin{figure}[hbt!]
    \centering
    \includegraphics[width=\textwidth]{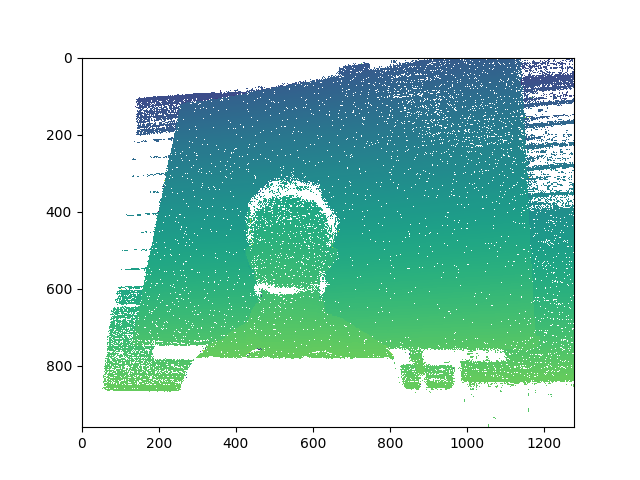}
    \caption{Decoding in y-direction using graycoded SL}
    \label{fig:y}
\end{figure}

Results of decoding the gray-coded structured light are shown in figure \ref{fig:scene}, figure \ref{fig:x} and figure \ref{fig:y}. Figure \ref{fig:scene} represents the original scene. In figure \ref{fig:x} is the decoding done in x-direction. In figure \ref{fig:y} is the decoding done in y-direction.

\section{Results after Point Cloud Registration}
\begin{figure}[hbt!]
    \centering
    \includegraphics[width=\textwidth]{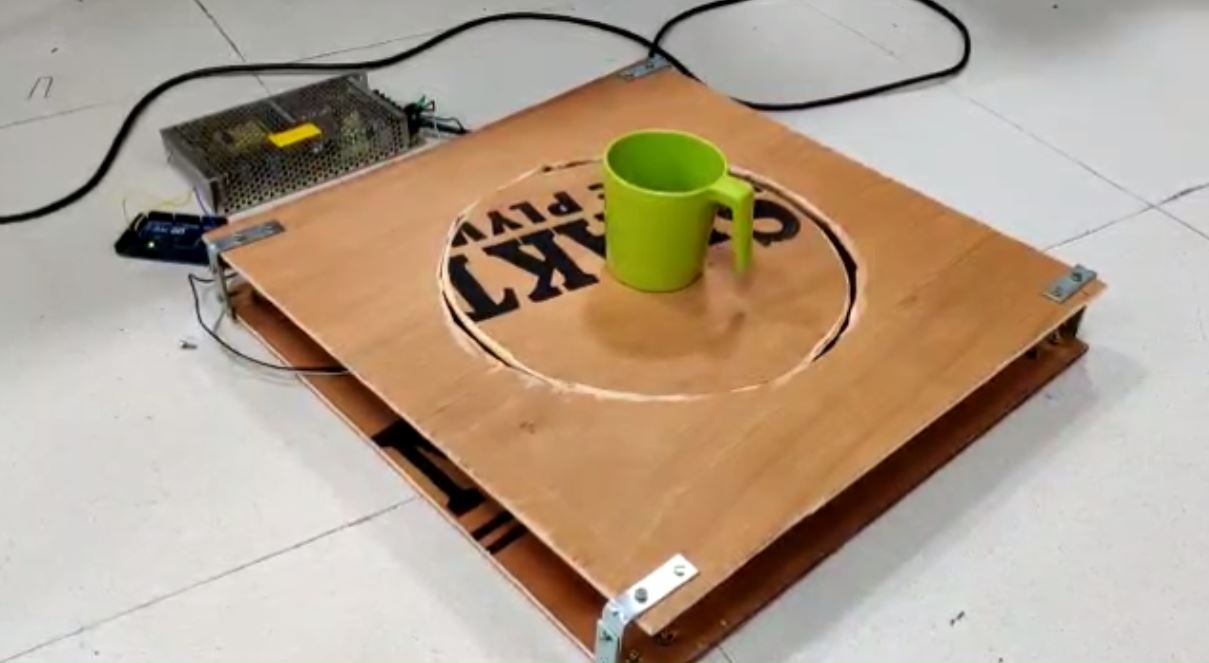}
    \caption{Cup and turntable assembly}
    \label{fig:scene2}
\end{figure}

\begin{figure}[hbt!]
    \centering
    \includegraphics[width=\textwidth]{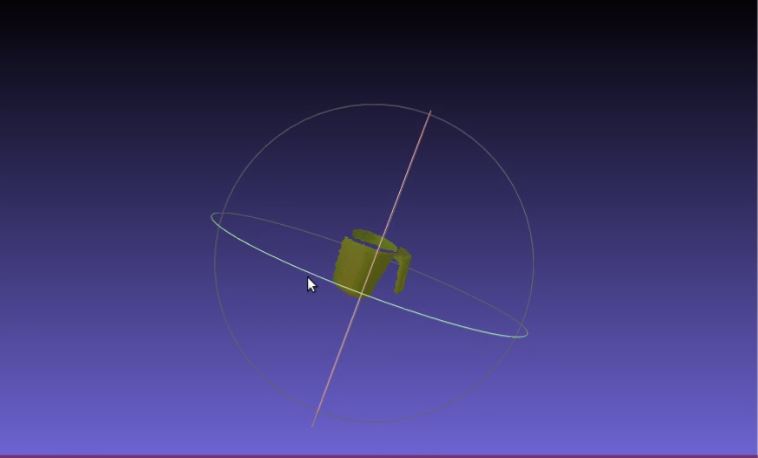}
    \caption{Reconstructed Cup with 10 point-clouds}
    \label{fig:reconstructed}
\end{figure}

10 images of the cup were taken after keeping it on turn-table. The original feature-less cup is shown in figure \ref{fig:scene2}. Final stitched point cloud of cup is shown in figure \ref{fig:reconstructed} .

\section{Point Clouds}
\begin{figure}[hbt!]
    \centering
    \includegraphics[width=\textwidth]{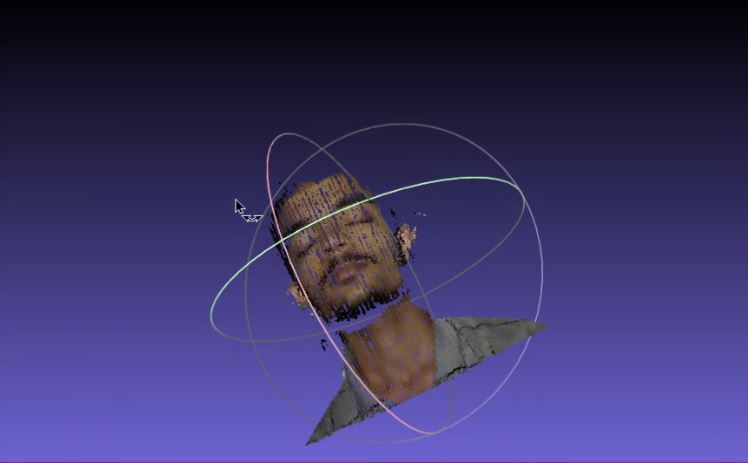}
    \caption{Face}
    \label{fig:face}
\end{figure}

\begin{figure}[hbt!]
    \centering
    \includegraphics[width=\textwidth]{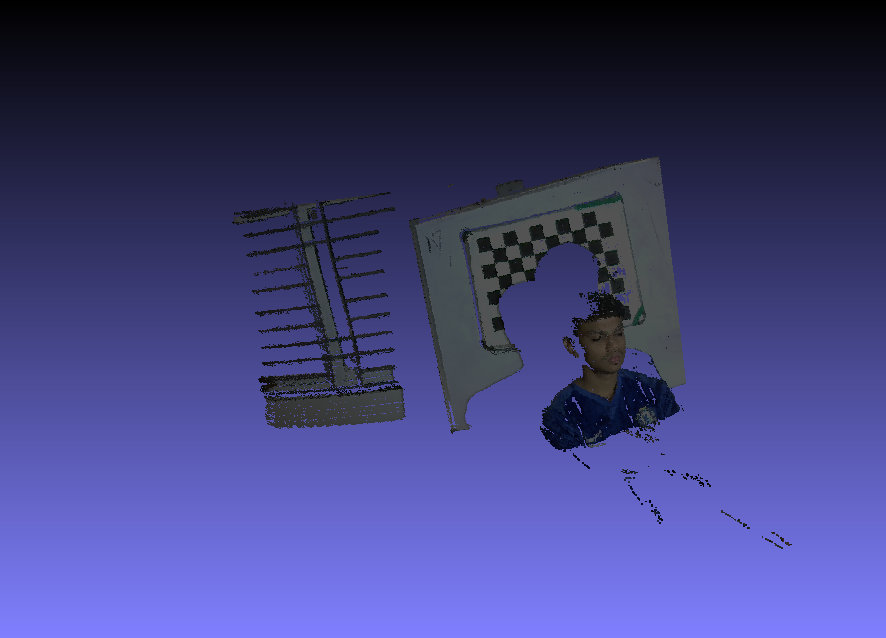}
    \caption{Random scene from ivlabs}
    \label{fig:ivlabs}
\end{figure}

Miscellaneous point clouds which are reconstructed using this algorithm. Figure \ref{fig:face} shows accuracy of face reconstruction from the proposed algorithm. Figure \ref{fig:ivlabs} shows the reconstruction of a random scene.

\section{Conclusion and Future Work}
We have been able to successfully implement a 3D scanner which uses gray-coded structured light. From the point clouds it is evident that it can suitably reconstruct featureless and complex
geometries. However, accuracy estimation of those point-clouds is yet to be done. 

There are many improvements that can be done in future to the existing pipeline. These are as follows:
\begin{itemize}
    \item Calibration of turntable in order to improve the accuracy of point cloud stitching algorithm. At present these are generated from the signals given by Dynamixel.
    \item Better mechanical setup so that the object to be reconstructed is always in focus. At present, it is done manually.
    \item Use of IR Projector instead of visible light, to make it robust against scene-illumination. At present we are estimating direct and indirect light. 
    \item Use of X-rays in order to reconstruct hollow cavities.
\end{itemize}
\printbibliography
\end{document}